\documentclass[10pt,twocolumn,letterpaper]{article}

\usepackage{cvpr}      % To produce the REVIEW version
\usepackage{graphicx}
\usepackage{amsmath}
\usepackage{amssymb}
\usepackage{booktabs}
\usepackage{multirow}
\usepackage{mathptmx}
\usepackage[mathcal]{eucal}
\usepackage{bm} % for bold math symbols

\usepackage[pagebackref,breaklinks,colorlinks]{hyperref}

\usepackage[capitalize]{cleveref}
\crefname{section}{Sec.}{Secs.}
\Crefname{section}{Section}{Sections}
\Crefname{table}{Table}{Tables}
\crefname{table}{Tab.}{Tabs.}

\def\cvprPaperID{*****} % *** Enter the CVPR Paper ID here
\def\confName{CVPR}
\def\confYear{2025}

\begin{document}

\title{FPC-Net: Revisiting SuperPoint with Descriptor-Free Keypoint Detection via Feature Pyramids and Consistency-Based Implicit Matching}

\author{Ionu\c{t} Grigore, Călin-Adrian Popa \\
Department of Computers and Information Technology\\
Politehnica University of Timi\c{s}oara\\
{\tt\small ionut.grigore-atimut@student.upt.ro}
\and
Claudiu Leoveanu-Condrei\\
ExtensityAI\\
{\tt\small leo@extensity.ai}
}
\maketitle

%%%%%%%%% ABSTRACT
\begin{abstract}
The extraction and matching of interest points are fundamental to many geometric computer vision tasks. Traditionally, matching is performed by assigning descriptors to interest points and identifying correspondences based on descriptor similarity. This work introduces a technique where interest points are inherently associated during detection, eliminating the need for computing, storing, transmitting, or matching descriptors. Although the matching accuracy is marginally lower than that of conventional approaches, our method completely eliminates the need for descriptors, leading to a drastic reduction in memory usage for localization systems. We assess its effectiveness by comparing it against both classical handcrafted methods and modern learned approaches. Code is available at \url{https://github.com/ionut-grigore99/FPC-Net}.
\end{abstract}

%%%%%%%%% BODY TEXT
\section{Introduction}
\label{sec:intro}

Feature extraction and matching play a fundamental role in applications such as visual localization \cite{sattler2018benchmarking}, structure-from-motion \cite{schonberger2016structure}, and visual odometry \cite{nister2006visual}. These applications rely heavily on accurately identifying salient points across different views and establishing reliable correspondences. The quality of keypoints and their associated descriptors directly influences downstream tasks such as pose estimation, mapping, and 3D reconstruction. Thus, designing reliable and efficient keypoint detectors remains a central challenge.

Traditional methods like SIFT \cite{lowe2004distinctive}, SURF \cite{bay2006surf}, and ORB \cite{rublee2011orb} introduced handcrafted detectors and binary descriptors that performed well under moderate transformations. However, their performance degrades in challenging scenarios involving large viewpoint or illumination changes. Deep learning-based methods such as LIFT \cite{yi2016lift}, SuperPoint \cite{detone2018superpoint}, and R2D2 \cite{revaud2019r2d2} have significantly improved detection robustness by learning keypoints and descriptors end-to-end from data. Despite these advancements, many models remain computationally expensive or lack sufficient flexibility to adapt to real-time or resource-constrained settings.

\begin{figure}[t]
  \centering
   \includegraphics[width=1\linewidth]{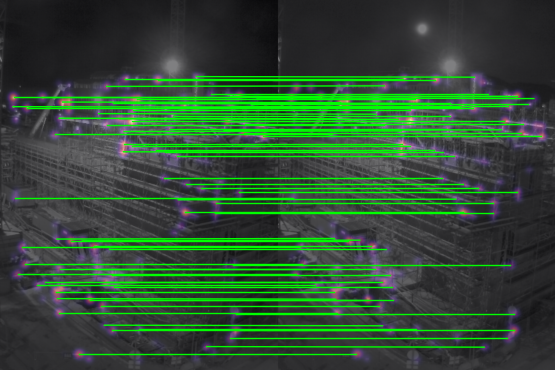}
   \caption{Qualitative matching example using our method. The figure shows reliable correspondences between two challenging views with significant lighting and perspective changes. Our approach obtains spatially consistent and semantically meaningful keypoints.}
   \label{fig:FPC-Net_matches}
\end{figure}

In this work, we propose an alternative training paradigm using a student-teacher framework to enhance keypoint detection. We leverage SuperPoint \cite{detone2018superpoint} as the teacher to provide structured supervision for a MobileNetV3 \cite{howard2019searching}-based student network, enhanced with a Feature Pyramid Network \cite{lin2017feature} to capture multi-scale features.

Our approach includes a two-stage training strategy to improve robustness and spatial consistency. In the second stage, we introduce a consistency loss and smooth target masks generated with LightGlue \cite{lindenberger2023lightglue}, framing keypoint detection as both a regression and classification problem.

Overall, our method achieves accurate and efficient keypoint detection, offering a strong tradeoff between performance and computational cost. We evaluate our model against both classical handcrafted methods and modern learned alternatives, demonstrating its effectiveness in challenging scenarios.

Our main contributions are as follows:

\begin{itemize}
    \item We propose a novel student-teacher framework for keypoint detection, where SuperPoint provides structured supervision to a lightweight student network. To increase stability under homographic transformations, we introduce a consistency loss that can be framed as either a regression or classification objective.
\end{itemize}

\begin{itemize}
    \item We design an efficient architecture based on MobileNetV3 enhanced with a Feature Pyramid Network (FPN) to improve multi-scale spatial representation while maintaining low computational cost.
\end{itemize}

\begin{itemize}
    \item We introduce a two-stage training strategy that first learns strong feature representations and then refines them using label smoothing and Gaussian-filtered masks to improve spatial consistency.
\end{itemize}

\begin{itemize}
    \item We demonstrate that our approach produces consistent, high-quality keypoint heatmaps and achieves competitive results on benchmark datasets while remaining efficient and scalable.
\end{itemize}

%------------------------------------------------------------------------
\section{Related work}
\label{sec:formatting}

The field of feature matching in computer vision has evolved significantly, shaped by both algorithmic constraints and the increasing demand for scalability. Early approaches prioritized sparse matching strategies \cite{sattler2018benchmarking, taira2018inloc, zhou2016evaluating, detone2018superpoint, dusmanu2019d2, ono2018lf, revaud2019r2d2,yi2016lift} due to the computational infeasibility of dense correspondence estimation \cite{farneback2003two, horn1981determining, rocco2018neighbourhood, rocco2020efficient, li2020dual, sun2021loftr, chen2022aspanformer, wang2022matchformer, choy2016universal, fathy2018hierarchical, savinov2017matching, schonberger2018semantic}. Pioneering detectors focused on identifying isolated, distinctive pixels—those that stand out from their neighbors and are more likely to be uniquely identified in other views. Over time, various detectors emerged, ranging from gradient-based techniques \cite{harris1988combined, shi1994good} to fast corner and blob detectors \cite{rosten2006machine}. These methods generally assign a score to each pixel and retain only the top responses, often applying non-maximum suppression to ensure spatial diversity.

Once such interest points are extracted, the challenge shifts to establishing reliable matches across views. Initial strategies used raw image patches, but these were quickly replaced by descriptors—compact representations designed to be invariant to certain transformations, such as minor viewpoint or illumination changes. Traditional descriptors like SIFT \cite{lowe2004distinctive} and HoG \cite{dalal2005histograms} encoded gradient information, while binary descriptors emerged as computationally lighter alternatives \cite{calonder2011brief, leutenegger2011brisk, rublee2011orb}.

Despite their success, handcrafted descriptors struggle with large-scale transformations like rotations or scale changes. This limitation motivated the use of multi-scale detection schemes \cite{leutenegger2011brisk, mikolajczyk2004scale} and local orientation estimation \cite{lowe2004distinctive, rublee2011orb} to improve stability under geometric transformations. However, these pipelines—comprising independent detection, description, and matching stages—often required delicate tuning and lacked adaptability to new domains.

The rise of deep learning reshaped this landscape. Convolutional neural networks (CNNs) proved capable of learning not just descriptors but entire detection pipelines. CNN-based detectors can be trained for invariance across geometric and photometric changes \cite{lenc2016learning}, yielding sharper and more stable interest point predictions \cite{zhang2018learning}. Moreover, CNNs can learn to rank keypoints consistently \cite{savinov2017quad} and provide high-quality outputs that outperform traditional handcrafted metrics in various settings. More recently, transformer architectures and attention-based models \cite{lindenberger2023lightglue, sun2021loftr} have further advanced this field by enabling global context modeling for feature representation and matching. These methods can capture long-range dependencies and enhance stability in complex scenes with repetitive structures or large viewpoint changes. In parallel, graph neural networks (GNNs) \cite{sarlin2020superglue} have been employed to reason about spatial relationships between keypoints, refining matches through structured attention mechanisms. These innovations, often layered on top of CNN-based feature extractors, have led to more accurate and context-aware matching pipelines.

Architectures, such as SuperPoint \cite{detone2018superpoint} and SuperGlue \cite{sarlin2020superglue}, embraced full deep pipelines by learning both keypoint detection and description in a unified framework. SuperPoint, for example, shares encoder layers between detection and description heads and trains via a synthetic-to-real transfer approach with homographic warping. SuperGlue, in turn, builds on top of SuperPoint keypoints and descriptors, using an attention-based graph neural network to perform context-aware matching between images—while still relying on explicit descriptors for correspondence estimation.

While these models have brought end-to-end learning to feature extraction, they still maintain an explicit notion of descriptors. Our approach challenges this paradigm by discarding traditional descriptors entirely. Instead, we introduce a unified architecture where the notion of a keypoint is intrinsically tied to the output activation channels of the network itself. This implicit representation enables matching without needing to extract or compare explicit descriptor vectors, reducing computational and memory overhead.

In doing so, we not only simplify the matching pipeline but also open the door to lightweight, communication-efficient systems. Unlike previous descriptor compression efforts—which rely on quantization or vocabulary encoding \cite{lynen2015get,tardioli2015visual}—our method inherently encodes correspondence information in the detection process, eliminating the need for descriptors altogether.

%------------------------------------------------------------------------
\section{Method}

\begin{figure*}[t]
  \centering
   \includegraphics[width=\linewidth]{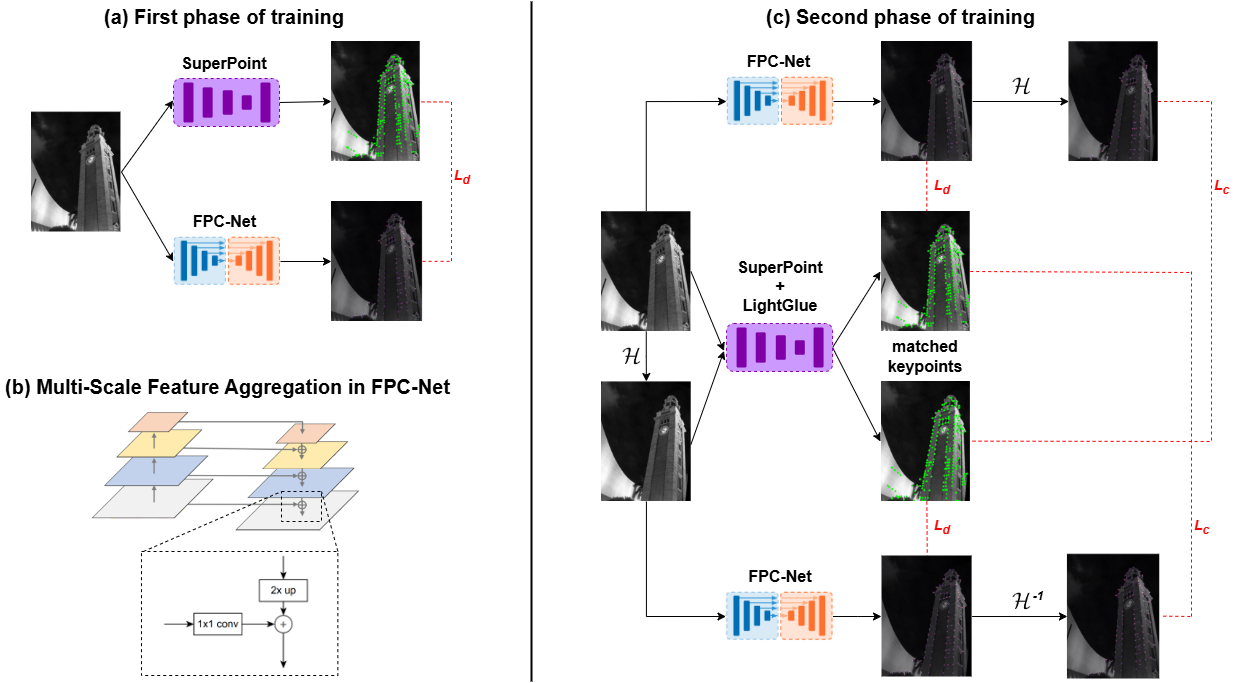}
   \caption{
    \textbf{Overview of the proposed two-stage training framework.}
    \textbf{(a)} In the first phase, we use SuperPoint to generate pseudo-ground-truth keypoint masks, which supervise the FPC-Net detector through a sigmoid focal loss $L_d$.
    \textbf{(b)} FPC-Net is built on a MobileNetV3 backbone with a Feature Pyramid Network (FPN) for multi-scale feature aggregation, combining representations from four stages via bicubic upsampling and 1$\times$1 convolutions.
    \textbf{(c)} In the second phase, LightGlue is used to obtain matched keypoints between original and warped images. Gaussian-filtered masks derived from these matches are used to supervise FPC-Net predictions under homographies $H$ and $H^{-1}$, through a consistency loss $L_c$. The focal loss $L_d$ remains active in both phases to encourage sharp and discriminative heatmap predictions.
   }
   \label{fig:overview}
\end{figure*}

Our approach utilizes SuperPoint \cite{detone2018superpoint} as a teacher model to generate keypoint labels, enabling a more structured learning process for a student network. The backbone of our model is MobileNetV3 \cite{howard2019searching}, a lightweight and efficient feature extractor, which we augment with a Feature Pyramid Network \cite{lin2017feature} (FPN) at four strategic locations to enhance multi-scale feature representation. The feature maps undergo a streamlined aggregation process—bicubic upscaling, summation, and dimensionality reduction via a convolutional layer—before being interpolated to a VGA-resolution heatmap. We are leveraging keypoints to guide the pretrained backbone's feature maps to concentrate on regions pertinent to keypoint detection. The resulting model yields heatmaps consistent across frames and more responsive to keypoint-like regions. Heatmaps offer a broader applicability beyond keypoints alone, allowing for experiments in deriving various features such as lines or shapes.

\begin{table*}[t]
\centering
\label{tab:repeatability}
\resizebox{0.8\textwidth}{!}{%
\begin{tabular}{@{}lccccccccc@{}}
\toprule
\multirow{2}{*}{\textbf{Method}} & \multicolumn{3}{c}{\textbf{all}} & \multicolumn{3}{c}{\textbf{i}} & \multicolumn{3}{c}{\textbf{v}} \\
\cmidrule(lr){2-4} \cmidrule(lr){5-7} \cmidrule(lr){8-10}
 & eps=1 & eps=3 & eps=8 & eps=1 & eps=3 & eps=8 & eps=1 & eps=3 & eps=8 \\
\midrule
FPC-Net (ours) & \textbf{0.46} & \textbf{0.59} & 0.67 & \textbf{0.43} & \textbf{0.51} & 0.55 & 0.48 & 0.67 & 0.79 \\

SuperPoint & 0.31 & 0.53 & 0.65 & 0.33 & \textbf{0.51} & 0.62 & 0.29 & 0.55 & 0.68 \\
Shi & 0.27 & 0.44 & 0.59 & 0.28 & 0.41 & 0.56 & 0.25 & 0.47 & 0.62 \\
Harris & 0.45 & \textbf{0.59} & 0.68 & 0.40 & 0.48 & 0.57 & \textbf{0.50} & \textbf{0.70} & 0.79 \\
FAST & 0.31 & 0.55 & \textbf{0.74} & 0.32 & 0.48 & \textbf{0.68} & 0.29 & 0.61 & \textbf{0.80} \\
SIFT & 0.27 & 0.46 & 0.70 & 0.27 & 0.41 & 0.63 & 0.27 & 0.52 & 0.77 \\
\bottomrule
\end{tabular}%
}
\caption{\textbf{HPatches Detector Repeatability}. Repeatability is measured as the fraction of correctly re-detected keypoints under different levels of geometric tolerance (eps = 1, 3, 8 pixels). Results are reported separately for illumination changes (i), viewpoint changes (v), and the full set (all) of HPatches image sequences.}
\end{table*}

\begin{table*}[t]
\centering
\label{tab:homography estimation}
\resizebox{0.95\textwidth}{!}{%
\begin{tabular}{@{}lccccccccccccc@{}}
\toprule
\multirow{2}{*}{\textbf{Method}} & \multirow{2}{*}{\textbf{Time (ms)}} & \multirow{2}{*}{\textbf{Size (MB)}} & \multicolumn{3}{c}{\textbf{all}} & \multicolumn{3}{c}{\textbf{i}} & \multicolumn{3}{c}{\textbf{v}} \\
\cmidrule(lr){4-6} \cmidrule(lr){7-9} \cmidrule(lr){10-12}
 & & & eps=1 & eps=3 & eps=8 & eps=1 & eps=3 & eps=8 & eps=1 & eps=3 & eps=8 \\
\midrule
FPC-Net (ours) & \textbf{8} & \textbf{0} & \textbf{0.54} & 0.74 & 0.84 & \textbf{0.63} & \textbf{0.88} & \textbf{0.97} & \textbf{0.44} & 0.60 & 0.70 \\
SuperPoint     & 200 & 614 & 0.36 & 0.75 & \textbf{0.93} & 0.46 & 0.84 & \textbf{0.97} & 0.26 & 0.66 & \textbf{0.89} \\
BRISK           & 78 & 153 & 0.31 & 0.64 & 0.78 & 0.38 & 0.67 & 0.76 & 0.24 & 0.62 & 0.80 \\
SIFT           & 40 & 307.2 & 0.44 & \textbf{0.78} & 0.89 & 0.51 & 0.81 & 0.88 & 0.37 & \textbf{0.74} & \textbf{0.89} \\
ORB            & 20 & 76.8 & 0.17 & 0.43 & 0.58 & 0.28 & 0.47 & 0.57 & 0.07 & 0.41 & 0.60 \\
\bottomrule
\end{tabular}%
}
\caption{\textbf{HPatches Homography Estimation.} A homography is considered correct if the average projection error of the warped image corners is below \textit{eps} pixels. We report results for illumination (\textbf{i}), viewpoint (\textbf{v}), and the full set (\textbf{all}), along with \textbf{total runtime} (ms per pair) and \textbf{descriptor size} (MB per pair).}
\end{table*}

We employ a two-stage training strategy to refine keypoint detection and improve spatial consistency. The first stage fine-tunes the backbone alongside the FPN to develop a strong representation of "keypointness", enriching feature maps with more meaningful spatial information, as illustrated in \cref{fig:overview} (a).  To supervise the detection head, we employ the sigmoid focal loss \cite{lin2017focal}—a variant of the standard cross-entropy loss that mitigates the imbalance between foreground and background pixels. It down-weights well-classified examples and focuses training on hard, ambiguous regions. We use its default parameters $\gamma=2.0$ and $\alpha=0.25$.

\begin{figure*}[t]
  \centering
   \includegraphics[width=\linewidth]{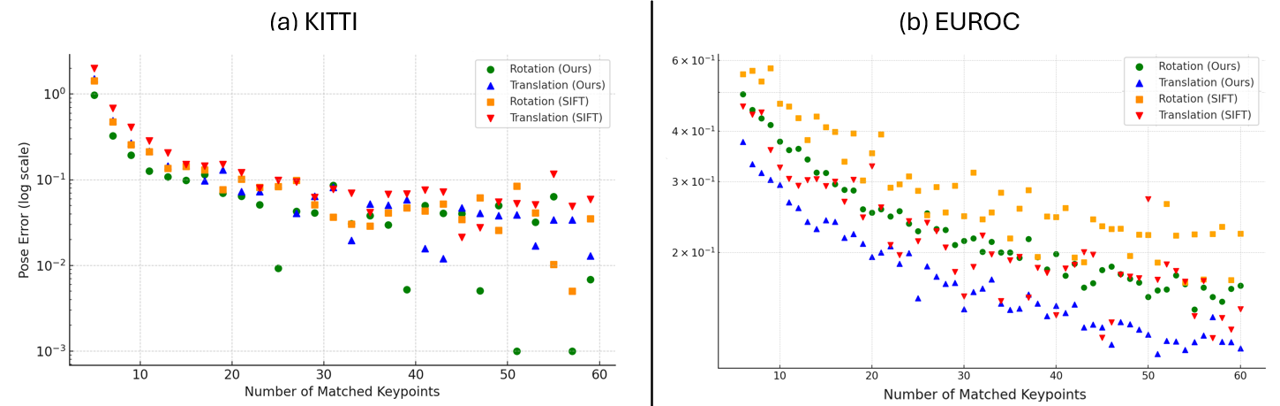}
   \caption{
    \textbf{Pose estimation error versus number of matched keypoints for (a) KITTI and (b) EuRoC datasets.} We report both rotation and translation errors in log scale for our method and SIFT. For both datasets, our method consistently achieves lower or comparable pose errors, particularly in low-keypoint regimes. This demonstrates its ability to produce more geometrically informative matches, especially when only a few correspondences are available. Notably, the performance gap is most significant below 30 keypoints, where our method yields accurate pose estimates despite the sparse input.
    }
   \label{fig:pose_estimation_accuracy_kitti_euroc}
\end{figure*}

The sigmoid focal loss used in our training is defined as:

\begin{align}
\mathcal{L}_{\text{focal}} =\; & -\alpha (1 - \sigma(\hat{y}))^\gamma y \log(\sigma(\hat{y})) \notag \\
& - (1 - \alpha)\, \sigma(\hat{y})^\gamma (1 - y) \log(1 - \sigma(\hat{y}))
\label{eq:focal}
\end{align}

where $\hat{y}$ is the predicted heatmap value, $y \in \{0, 1\}$ is the ground truth keypoint mask, $\sigma(\cdot)$ denotes the sigmoid function, and $\alpha$, $\gamma$ are the balancing parameters.

The second stage as shown in \cref{fig:overview} (c) builds upon this foundation, introducing label smoothing and Gaussian-filtered target masks to encourage spatial awareness and uncertainty modeling. These masks are derived using LightGlue \cite{lindenberger2023lightglue}, ensuring that keypoints remain consistent between original and warped images. The sigmoid focal loss remains active during this phase to continue supervising the detector head with hard-negative mining and class imbalance mitigation. We augment the second phase by incorporating a consistency loss alongside the active sigmoid focal loss, aiming to enhance invariance to homographic transformations. For this, we experiment with the Huber loss, which offers a balance between L1 and L2 penalties, allowing smoother gradient propagation while handling outliers more effectively.

Additionally, we explore an alternative approach by formulating keypoint detection as a fine-tuning classification problem, leveraging KL-divergence to treat the image as a probability distribution. This perspective shifts the focus towards learning a structured distribution of keypoint locations rather than explicit coordinate regression. However, because certain image regions may become occluded or invalid after warping (e.g., moving outside the image bounds), we apply a binary validity mask that excludes these regions from contributing to the loss. This ensures that consistency is only enforced where pixel-level correspondence is well-defined.

The consistency loss can be mathematically defined as follows:
\begin{align}
\mathcal{L}_{\text{C}_{\mathrm{reg}}} =\; & \text{Huber}\left[\sigma(pH), m'\right] \notag \\
& + \text{Huber}\left[\sigma(p'H^{-1}), m\right]
\label{eq:regression}
\end{align}

\begin{align}
\mathcal{L}_{\text{C}_{\mathrm{clf}}} =\; & \text{KL}\left[\log(\mathcal{S}(pH)), \mathcal{S}(m')\right] \notag \\
& + \text{KL}\left[\log(\mathcal{S}(p'H^{-1})), \mathcal{S}(m)\right]
\label{eq:classification}
\end{align}

where $p$, $p'$ denote the predictions for the original and warped images, $H$ is a homography, $m$, $m'$ are the corresponding ground truth masks, $\sigma$ is the sigmoid function, and $\mathcal{S}$ is the softmax function.

%------------------------------------------------------------------------
\section{Experiments}
In our experiments, we train the FPC-Net model on MS-COCO images and evaluate its performance on the HPatches dataset \cite{balntas2017hpatches}, the initial sequence (00) from the KITTI autonomous driving dataset \cite{geiger2013vision}, and sequence V1\_01 from the EuRoC MAV dataset \cite{burri2016euroc}, which captures indoor drone navigation. 

HPatches consists of 116 scenes and 696 unique images, with the first 57 scenes featuring significant illumination changes and the remaining 59 showcasing large viewpoint variations.

To construct evaluation pairs for the KITTI and EuRoC datasets, we randomly select 100 image pairs using the sampling method detailed below. Although both datasets offer stereo sequences, our experiments are limited to images from the left camera only.

The procedure for generating pairs from an image sequence is as follows: starting from a given reference frame, we densely initialize 2D points and track them forward using the KLT algorithm across subsequent frames. A pair is then formed between the reference frame and a randomly chosen later frame in which at least a fraction $\alpha$ of the original points remain successfully tracked. The parameter $\alpha$ effectively controls the minimum required scene overlap between the two views. This strategy offers the advantage of being agnostic to camera calibration, while still ensuring a reliable level of shared scene content. In our setup, we use a stricter $\alpha = 0.5$ when generating  evaluation pairs to ensure higher overlap during testing.

%-------------------------------------------------------------------------
\subsection{Keypoint Repeatability on HPatches}

To assess FPC-Net’s interest point detection capabilities, we measure repeatability on the HPatches dataset. We compare its performance against the SuperPoint model, as well as FAST \cite{rosten2006machine}, Harris \cite{harris1988combined}, and Shi \cite{shi1994good}. 

To evaluate the performance of an interest point detector across image pairs, we measure repeatability, defined as the the proportion of keypoints detected in one image that also appear in the corresponding warped image within a given pixel threshold.
All evaluations are performed at a resolution of 640 × 480, with 300 keypoints detected per image. For SuperPoint, Non-Maximum Suppression (NMS) with a radius of 4 pixels is applied to select the top responses. The correct match distance threshold is varied across 1, 3, and 8 pixels to provide a comprehensive assessment under different levels of geometric tolerance.

Let us denote the set of detected keypoints in the first image as $\{\bm{p}_m\}_{m=1}^{M}$, and those in the second image as $\{\bm{q}_n\}_{n=1}^{N}$. We define a tolerance radius $\epsilon$ to determine whether a keypoint match is valid. A point $\bm{p}_m$ is considered successfully repeated if it has at least one corresponding point $\bm{q}_n$ within that radius. Mathematically, the correctness indicator for a point $\bm{p}_m$ is given by:

\begin{equation}
\text{Match}(\bm{p}_m) = \left( \min_{n \in \{1, \ldots, N\}} \|\bm{p}_m - \bm{q}_n\| \leq \epsilon \right)
\end{equation}

Likewise, we apply the same rule symmetrically to each point $\bm{q}_n$ from the second image. The final repeatability score is then computed as the average fraction of matched keypoints from both sets:

{\small
\begin{equation}
\text{Repeatability} = \frac{1}{M + N} \left( \sum_{m=1}^{M} \text{Match}(\bm{p}_m) + \sum_{n=1}^{N} \text{Match}(\bm{q}_n) \right)
\end{equation}
}

This value reflects the fraction of points that are consistently detected between the two images.

In summary, applying the consistency loss enhances repeatability, particularly under substantial viewpoint changes. The corresponding results are presented in Table 1 where we can see that FPC-Net outperforms existing methods in repeatability while remaining descriptor-free and computationally efficient.

%-------------------------------------------------------------------------
\subsection{Homography Estimation Accuracy on HPatches}

To evaluate FPC-Net’s matching performance, we assess its matching capabilities on the HPatches dataset. All experiments are conducted on images resized to a resolution of 480 × 640. Unlike traditional systems that output both keypoints and descriptors, FPC-Net produces a heatmap indicating the keypointness of each pixel, without providing descriptors. To enable homography estimation, we extract keypoints by applying quantile-based thresholding to the heatmap, selecting the most confident locations. We then perform descriptor-free matching by comparing the spatial proximity of detected points. Specifically, we match keypoints between image pairs based on nearest neighbor search in image coordinates, and estimate the homography using RANSAC via OpenCV’s findHomography() function. FPC-Net is compared against three established systems: BRISK \cite{leutenegger2011brisk}, SIFT \cite{lowe2004distinctive}, and ORB \cite{rublee2011orb}. 

To assess how well an algorithm estimates a homography between two images, we evaluate how accurately it maps known reference points rather than directly comparing the $3 \times 3$ matrix values—which can be unreliable due to varying matrix entry scales. Instead, we use a geometric validation method based on how well four canonical points (e.g., image corners) from one image are transformed to their expected locations in the other image.

Let the corner coordinates in the source image be denoted by $\mathbf{p}_1, \mathbf{p}_2, \mathbf{p}_3, \mathbf{p}_4$. The true homography matrix $\mathbf{H}_\text{gt}$ maps these to $\mathbf{p}'_1, \mathbf{p}'_2, \mathbf{p}'_3, \mathbf{p}'_4$ in the second image. Meanwhile, the predicted homography $\widehat{\mathbf{H}}$ gives the transformed points $\hat{\mathbf{p}}'_1, \hat{\mathbf{p}}'_2, \hat{\mathbf{p}}'_3, \hat{\mathbf{p}}'_4$.

We define the homography correctness over a batch of $K$ samples by measuring the average reprojection error over the four corners, and checking whether the mean distance is below a threshold $\epsilon$:

\begin{equation}
\text{Accuracy}_{\text{H}} = \frac{1}{K} \sum_{k=1}^{K}
\left(
\left( \frac{1}{4} \sum_{c=1}^{4} \left\| \mathbf{p}'^{(k)}_c - \hat{\mathbf{p}}'^{(k)}_c \right\| \right) \leq \epsilon
\right)
\end{equation}

Here, $\mathbf{p}'^{(k)}_c$ represents the $c$-th ground-truth corner in sample $k$, while $\hat{\mathbf{p}}'^{(k)}_c$ is the corresponding point predicted by the estimated homography. The threshold $\epsilon$ defines the maximum permissible average corner error for a match to be counted as correct.

Scores range between 0 and 1, with values closer to 1 indicating more accurate homography estimation.

The homography estimation results, shown in Table 2, indicate that FPC-Net achieves better accuracy and reliability, all while remaining descriptor-free and efficient.

%-------------------------------------------------------------------------
\subsection{Pose Estimation Accuracy}

To understand how detection and matching performance impacts downstream pose estimation, we compare the relative camera pose recovered using our method versus SIFT, with respect to the ground truth on KITTI and EuRoC. For each image pair, we compute 3D point correspondences via stereo triangulation from interest point matches (left-right camera) and estimate pose using P3P \cite{gao2003complete} within a RANSAC loop \cite{fischler1981random}. The estimated poses are then compared to ground truth using rotation error (measured as geodesic distance in angle-axis representation) and translation error (measured in Euclidean distance). In \cref{fig:pose_estimation_accuracy_kitti_euroc} (a), we plot these errors against the number of inlier correspondences for each image pair in KITTI. The corresponding results for EuRoC are provided in the \cref{fig:pose_estimation_accuracy_kitti_euroc} (b).

%-------------------------------------------------------------------------
\subsection{Other experiments}
\begin{figure}[t]
  \centering
   \includegraphics[width=1\linewidth]{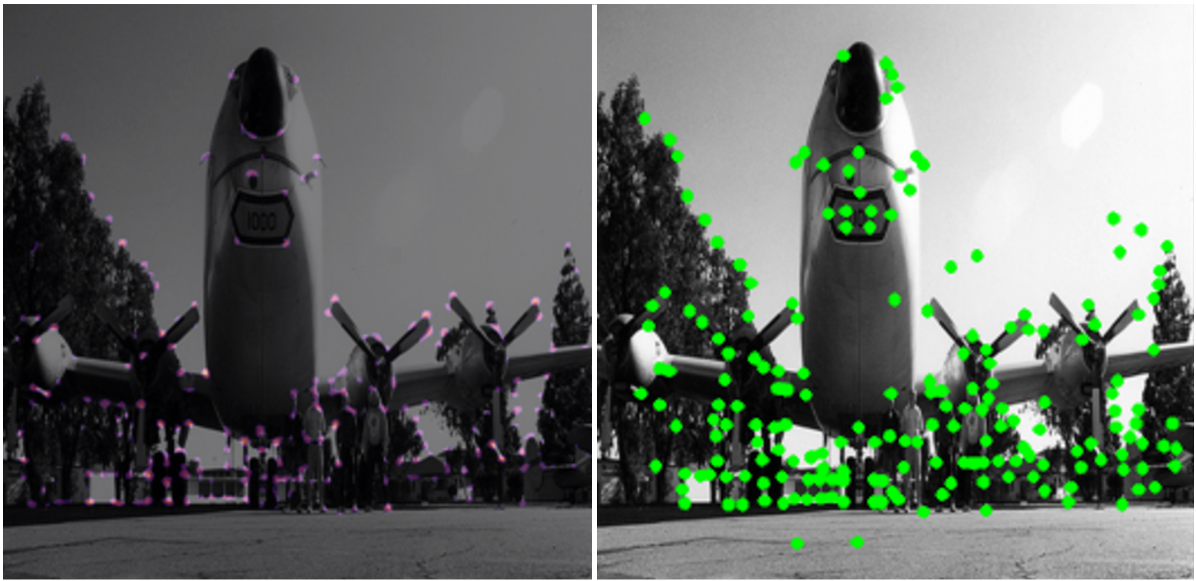}
   \caption{Comparison of keypoint detections between our network (left) and SuperPoint+LightGlue (right). Our method outputs clean, sparse, and semantically meaningful detections focused on salient object parts (e.g., aircraft body and propellers), while SuperPoint+LightGlue produces dense and redundant detections, particularly in texture-heavy regions like the ground and foliage.}
   \label{fig:FPC-Net_cleaner_predictions}
\end{figure}

To further assess the behavior of our network, we present a qualitative comparison between its predicted keypoints and those generated by the SuperPoint+LightGlue pipeline. As illustrated in \cref{fig:FPC-Net_cleaner_predictions}, our model produces cleaner, more spatially structured keypoints, with improved semantic coverage of salient object regions. In contrast, SuperPoint+LightGlue tends to produce dense and often redundant keypoints, particularly in texture-rich areas such as the ground or tree foliage.

These dense detections require non-maximum suppression (NMS) to achieve a usable sparse set, whereas our method directly outputs sharp heatmap peaks that reflect high confidence in semantically meaningful areas. This demonstrates the ability of our network to learn a more interpretable and focused keypoint distribution, which simplifies downstream processing and matching.

\begin{figure}[t]
  \centering
   \includegraphics[width=1\linewidth]{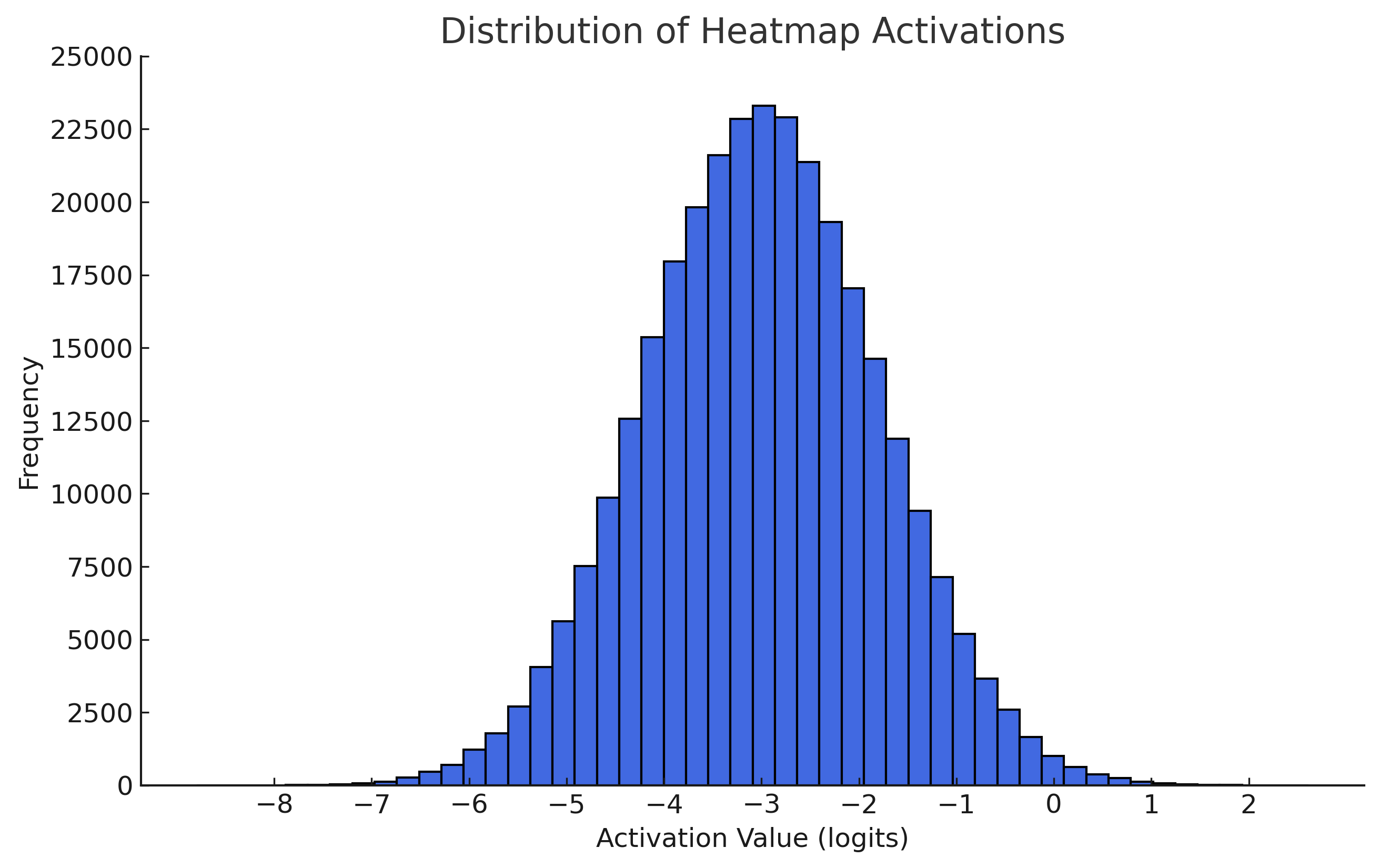}
   \caption{Distribution of raw heatmap activation values before sigmoid transformation. Most activations are centered around $-3$, suppressing low-confidence regions. Only a few high-activation peaks contribute to keypoint predictions after sigmoid, promoting sparsity and precision.}
   \label{fig:histogram_activations}
\end{figure}

To further support this observation, we visualize the distribution of heatmap activations in \cref{fig:histogram_activations}. The majority of activation values are sharply concentrated around $-3$, which after applying the sigmoid function corresponds to a very low confidence (i.e., $\sigma(-3) \approx 0.047$), effectively suppressing irrelevant regions. Only a small fraction of the activations exceed zero, indicating confident peaks for true keypoints. This distribution confirms that our network learns to produce sparse and confident detections, and that the sigmoid-based activation output acts as an implicit confidence filter.

%-------------------------------------------------------------------------
\subsection{Implementation Details}

All training is conducted using PyTorch \cite{paszke1912imperative} with a mini-batch size of 8 and the ADAM optimizer, configured with its default parameters (learning rate = 0.001, beta = (0.9, 0.999)). Experiments are run on a single NVIDIA Tesla V100-SXM2 GPU. The training follows a two-phase strategy: the first phase, focused on learning discriminative  keypoint representations, runs for 10 epochs, while the second phase, which incorporates spatial consistency and uncertainty modeling, is trained for an additional 6 epochs.

To strengthen the network’s adaptability to lighting, viewpoint, and local image distortions, we employ a diverse set of data augmentation techniques from the Albumentations library \cite{buslaev2020albumentations}. Specifically, we apply Glass blur, motion blur, defocus, and Gaussian noise to simulate sensor and camera noise. To account for geometric variability, we include perspective transforms, affine and piecewise affine deformations, and shift-scale-rotate augmentations. Additionally, we introduce random brightness and contrast changes to enhance illumination invariance. These augmentations are applied with high probability to encourage generalization under real-world image variations.

Our model adopts a lightweight yet expressive architecture, combining MobileNetV3-Small \cite{howard2019searching} as the backbone and a Feature Pyramid Network (FPN) \cite{lin2017feature} for multi-scale feature aggregation. We choose MobileNetV3 due to its efficient inverted residual bottlenecks and low computational footprint, making it suitable for real-time keypoint detection. From the MobileNetV3 feature extractor, we select intermediate outputs from four strategically chosen layers—indexed as {1, 2, 4, 12}—which correspond to progressively deeper semantic representations with increasing receptive fields.

These feature maps, with channel sizes {16, 24, 40, 576}, are fed into an FPN that projects them to a common embedding space of 128 channels using lateral 1×1 convolutions, followed by upsampling and top-down fusion. This hierarchical design, illustrated in \cref{fig:overview} (b) enhances the model’s ability to detect keypoints across scales by blending low-level texture and high-level context information. We then upsample the top-level FPN features (from layer 12) step-by-step, summing them with lower-level outputs (from layers 4, 2, and 1) using bicubic interpolation. The final aggregated feature map is passed through a 1×1 convolution and batch normalization layer to produce a single-channel heatmap that represents the likelihood of keypoint locations.

During inference, this heatmap is rescaled to the target resolution using bicubic interpolation and post-processed via quantile-based thresholding to retain only the most salient peaks. This design balances detection performance and computational efficiency, making the model deployable in real-time scenarios.

%------------------------------------------------------------------------
\section{Conclusions}

 This study introduces a technique where interest points are inherently associated during detection, eliminating the need for computing, storing, transmitting, or matching descriptors. The key novelty is that instead of relying on descriptors for matching, interest points are uniquely associated through a descriptor-free heatmap prediction, supervised using LightGlue matches and regularized with a KL-divergence consistency loss between warped and original images. Although our method does not rely on descriptors, it achieves matching accuracy on par with or better than conventional approaches, while drastically reducing memory usage in localization systems.
%%%%%%%%% REFERENCES
{\small
\bibliographystyle{ieee_fullname}
\bibliography{FPC-Net.bbl}

\begin{thebibliography}{10}\itemsep=-1pt

\bibitem{balntas2017hpatches}
Vassileios Balntas, Karel Lenc, Andrea Vedaldi, and Krystian Mikolajczyk.
\newblock Hpatches: A benchmark and evaluation of handcrafted and learned local descriptors.
\newblock In {\em Proceedings of the IEEE conference on computer vision and pattern recognition}, pages 5173--5182, 2017.

\bibitem{bay2006surf}
Herbert Bay, Tinne Tuytelaars, and Luc Van~Gool.
\newblock Surf: Speeded up robust features.
\newblock In {\em Computer Vision--ECCV 2006: 9th European Conference on Computer Vision, Graz, Austria, May 7-13, 2006. Proceedings, Part I 9}, pages 404--417. Springer, 2006.

\bibitem{burri2016euroc}
Michael Burri, Janosch Nikolic, Pascal Gohl, Thomas Schneider, Joern Rehder, Sammy Omari, Markus~W Achtelik, and Roland Siegwart.
\newblock The euroc micro aerial vehicle datasets.
\newblock {\em The International Journal of Robotics Research}, 35(10):1157--1163, 2016.

\bibitem{buslaev2020albumentations}
Alexander Buslaev, Vladimir~I Iglovikov, Eugene Khvedchenya, Alex Parinov, Mikhail Druzhinin, and Alexandr~A Kalinin.
\newblock Albumentations: fast and flexible image augmentations.
\newblock {\em Information}, 11(2):125, 2020.

\bibitem{calonder2011brief}
Michael Calonder, Vincent Lepetit, Mustafa Ozuysal, Tomasz Trzcinski, Christoph Strecha, and Pascal Fua.
\newblock Brief: Computing a local binary descriptor very fast.
\newblock {\em IEEE transactions on pattern analysis and machine intelligence}, 34(7):1281--1298, 2011.

\bibitem{chen2022aspanformer}
Hongkai Chen, Zixin Luo, Lei Zhou, Yurun Tian, Mingmin Zhen, Tian Fang, David Mckinnon, Yanghai Tsin, and Long Quan.
\newblock Aspanformer: Detector-free image matching with adaptive span transformer.
\newblock In {\em European Conference on Computer Vision}, pages 20--36. Springer, 2022.

\bibitem{choy2016universal}
Christopher~B Choy, JunYoung Gwak, Silvio Savarese, and Manmohan Chandraker.
\newblock Universal correspondence network.
\newblock {\em Advances in neural information processing systems}, 29, 2016.

\bibitem{dalal2005histograms}
Navneet Dalal and Bill Triggs.
\newblock Histograms of oriented gradients for human detection.
\newblock In {\em 2005 IEEE computer society conference on computer vision and pattern recognition (CVPR'05)}, volume~1, pages 886--893. Ieee, 2005.

\bibitem{detone2018superpoint}
Daniel DeTone, Tomasz Malisiewicz, and Andrew Rabinovich.
\newblock Superpoint: Self-supervised interest point detection and description.
\newblock In {\em Proceedings of the IEEE conference on computer vision and pattern recognition workshops}, pages 224--236, 2018.

\bibitem{dusmanu2019d2}
Mihai Dusmanu, Ignacio Rocco, Tomas Pajdla, Marc Pollefeys, Josef Sivic, Akihiko Torii, and Torsten Sattler.
\newblock D2-net: A trainable cnn for joint description and detection of local features.
\newblock In {\em Proceedings of the ieee/cvf conference on computer vision and pattern recognition}, pages 8092--8101, 2019.

\bibitem{farneback2003two}
Gunnar Farneb{\"a}ck.
\newblock Two-frame motion estimation based on polynomial expansion.
\newblock In {\em Image Analysis: 13th Scandinavian Conference, SCIA 2003 Halmstad, Sweden, June 29--July 2, 2003 Proceedings 13}, pages 363--370. Springer, 2003.

\bibitem{fathy2018hierarchical}
Mohammed~E Fathy, Quoc-Huy Tran, M~Zeeshan Zia, Paul Vernaza, and Manmohan Chandraker.
\newblock Hierarchical metric learning and matching for 2d and 3d geometric correspondences.
\newblock In {\em Proceedings of the european conference on computer vision (ECCV)}, pages 803--819, 2018.

\bibitem{fischler1981random}
Martin~A Fischler and Robert~C Bolles.
\newblock Random sample consensus: a paradigm for model fitting with applications to image analysis and automated cartography.
\newblock {\em Communications of the ACM}, 24(6):381--395, 1981.

\bibitem{gao2003complete}
Xiao-Shan Gao, Xiao-Rong Hou, Jianliang Tang, and Hang-Fei Cheng.
\newblock Complete solution classification for the perspective-three-point problem.
\newblock {\em IEEE transactions on pattern analysis and machine intelligence}, 25(8):930--943, 2003.

\bibitem{geiger2013vision}
Andreas Geiger, Philip Lenz, Christoph Stiller, and Raquel Urtasun.
\newblock Vision meets robotics: The kitti dataset.
\newblock {\em The international journal of robotics research}, 32(11):1231--1237, 2013.

\bibitem{harris1988combined}
Chris Harris, Mike Stephens, et~al.
\newblock A combined corner and edge detector.
\newblock In {\em Alvey vision conference}, volume~15, pages 10--5244. Citeseer, 1988.

\bibitem{horn1981determining}
Berthold~KP Horn and Brian~G Schunck.
\newblock Determining optical flow.
\newblock {\em Artificial intelligence}, 17(1-3):185--203, 1981.

\bibitem{howard2019searching}
Andrew Howard, Mark Sandler, Grace Chu, Liang-Chieh Chen, Bo Chen, Mingxing Tan, Weijun Wang, Yukun Zhu, Ruoming Pang, Vijay Vasudevan, et~al.
\newblock Searching for mobilenetv3.
\newblock In {\em Proceedings of the IEEE/CVF international conference on computer vision}, pages 1314--1324, 2019.

\bibitem{lenc2016learning}
Karel Lenc and Andrea Vedaldi.
\newblock Learning covariant feature detectors.
\newblock In {\em Computer Vision--ECCV 2016 Workshops: Amsterdam, The Netherlands, October 8-10 and 15-16, 2016, Proceedings, Part III 14}, pages 100--117. Springer, 2016.

\bibitem{leutenegger2011brisk}
Stefan Leutenegger, Margarita Chli, and Roland~Y Siegwart.
\newblock Brisk: Binary robust invariant scalable keypoints.
\newblock In {\em 2011 International conference on computer vision}, pages 2548--2555. Ieee, 2011.

\bibitem{li2020dual}
Xinghui Li, Kai Han, Shuda Li, and Victor Prisacariu.
\newblock Dual-resolution correspondence networks.
\newblock {\em Advances in Neural Information Processing Systems}, 33:17346--17357, 2020.

\bibitem{lin2017feature}
Tsung-Yi Lin, Piotr Doll{\'a}r, Ross Girshick, Kaiming He, Bharath Hariharan, and Serge Belongie.
\newblock Feature pyramid networks for object detection.
\newblock In {\em Proceedings of the IEEE conference on computer vision and pattern recognition}, pages 2117--2125, 2017.

\bibitem{lin2017focal}
Tsung-Yi Lin, Priya Goyal, Ross Girshick, Kaiming He, and Piotr Doll{\'a}r.
\newblock Focal loss for dense object detection.
\newblock In {\em Proceedings of the IEEE international conference on computer vision}, pages 2980--2988, 2017.

\bibitem{lindenberger2023lightglue}
Philipp Lindenberger, Paul-Edouard Sarlin, and Marc Pollefeys.
\newblock Lightglue: Local feature matching at light speed.
\newblock In {\em Proceedings of the IEEE/CVF International Conference on Computer Vision}, pages 17627--17638, 2023.

\bibitem{lowe2004distinctive}
David~G Lowe.
\newblock Distinctive image features from scale-invariant keypoints.
\newblock {\em International journal of computer vision}, 60:91--110, 2004.

\bibitem{lynen2015get}
Simon Lynen, Torsten Sattler, Michael Bosse, Joel~A Hesch, Marc Pollefeys, and Roland Siegwart.
\newblock Get out of my lab: Large-scale, real-time visual-inertial localization.
\newblock In {\em Robotics: Science and Systems}, volume~1, 2015.

\bibitem{mikolajczyk2004scale}
Krystian Mikolajczyk and Cordelia Schmid.
\newblock Scale \& affine invariant interest point detectors.
\newblock {\em International journal of computer vision}, 60:63--86, 2004.

\bibitem{nister2006visual}
David Nist{\'e}r, Oleg Naroditsky, and James Bergen.
\newblock Visual odometry for ground vehicle applications.
\newblock {\em Journal of Field Robotics}, 23(1):3--20, 2006.

\bibitem{ono2018lf}
Yuki Ono, Eduard Trulls, Pascal Fua, and Kwang~Moo Yi.
\newblock Lf-net: Learning local features from images.
\newblock {\em Advances in neural information processing systems}, 31, 2018.

\bibitem{paszke1912imperative}
Adam Paszke, Sam Gross, Francisco Massa, Adam Lerer, JP Bradbury, Gregory Chanan, Trevor Killeen, Zeming Lin, Natalia Gimelshein, Luca Antiga, et~al.
\newblock An imperative style, high-performance deep learning library.
\newblock {\em Adv. Neural Inf. Process. Syst}, 32:8026, 1912.

\bibitem{revaud2019r2d2}
Jerome Revaud, Cesar De~Souza, Martin Humenberger, and Philippe Weinzaepfel.
\newblock R2d2: Reliable and repeatable detector and descriptor.
\newblock {\em Advances in neural information processing systems}, 32, 2019.

\bibitem{rocco2020efficient}
Ignacio Rocco, Relja Arandjelovi{\'c}, and Josef Sivic.
\newblock Efficient neighbourhood consensus networks via submanifold sparse convolutions.
\newblock In {\em Computer vision--ECCV 2020: 16th European conference, Glasgow, UK, August 23--28, 2020, proceedings, part IX 16}, pages 605--621. Springer, 2020.

\bibitem{rocco2018neighbourhood}
Ignacio Rocco, Mircea Cimpoi, Relja Arandjelovi{\'c}, Akihiko Torii, Tomas Pajdla, and Josef Sivic.
\newblock Neighbourhood consensus networks.
\newblock {\em Advances in neural information processing systems}, 31, 2018.

\bibitem{rosten2006machine}
Edward Rosten and Tom Drummond.
\newblock Machine learning for high-speed corner detection.
\newblock In {\em Computer Vision--ECCV 2006: 9th European Conference on Computer Vision, Graz, Austria, May 7-13, 2006. Proceedings, Part I 9}, pages 430--443. Springer, 2006.

\bibitem{rublee2011orb}
Ethan Rublee, Vincent Rabaud, Kurt Konolige, and Gary Bradski.
\newblock Orb: An efficient alternative to sift or surf.
\newblock In {\em 2011 International conference on computer vision}, pages 2564--2571. Ieee, 2011.

\bibitem{sarlin2020superglue}
Paul-Edouard Sarlin, Daniel DeTone, Tomasz Malisiewicz, and Andrew Rabinovich.
\newblock Superglue: Learning feature matching with graph neural networks.
\newblock In {\em Proceedings of the IEEE/CVF conference on computer vision and pattern recognition}, pages 4938--4947, 2020.

\bibitem{sattler2018benchmarking}
Torsten Sattler, Will Maddern, Carl Toft, Akihiko Torii, Lars Hammarstrand, Erik Stenborg, Daniel Safari, Masatoshi Okutomi, Marc Pollefeys, Josef Sivic, et~al.
\newblock Benchmarking 6dof outdoor visual localization in changing conditions.
\newblock In {\em Proceedings of the IEEE conference on computer vision and pattern recognition}, pages 8601--8610, 2018.

\bibitem{savinov2017matching}
Nikolay Savinov, Lubor Ladicky, and Marc Pollefeys.
\newblock Matching neural paths: transfer from recognition to correspondence search.
\newblock {\em Advances in Neural Information Processing Systems}, 30, 2017.

\bibitem{savinov2017quad}
Nikolay Savinov, Akihito Seki, Lubor Ladicky, Torsten Sattler, and Marc Pollefeys.
\newblock Quad-networks: unsupervised learning to rank for interest point detection.
\newblock In {\em Proceedings of the IEEE conference on computer vision and pattern recognition}, pages 1822--1830, 2017.

\bibitem{schonberger2016structure}
Johannes~L Schonberger and Jan-Michael Frahm.
\newblock Structure-from-motion revisited.
\newblock In {\em Proceedings of the IEEE conference on computer vision and pattern recognition}, pages 4104--4113, 2016.

\bibitem{schonberger2018semantic}
Johannes~L Sch{\"o}nberger, Marc Pollefeys, Andreas Geiger, and Torsten Sattler.
\newblock Semantic visual localization.
\newblock In {\em Proceedings of the IEEE conference on computer vision and pattern recognition}, pages 6896--6906, 2018.

\bibitem{shi1994good}
Jianbo Shi et~al.
\newblock Good features to track.
\newblock In {\em 1994 Proceedings of IEEE conference on computer vision and pattern recognition}, pages 593--600. IEEE, 1994.

\bibitem{sun2021loftr}
Jiaming Sun, Zehong Shen, Yuang Wang, Hujun Bao, and Xiaowei Zhou.
\newblock Loftr: Detector-free local feature matching with transformers.
\newblock In {\em Proceedings of the IEEE/CVF conference on computer vision and pattern recognition}, pages 8922--8931, 2021.

\bibitem{taira2018inloc}
Hajime Taira, Masatoshi Okutomi, Torsten Sattler, Mircea Cimpoi, Marc Pollefeys, Josef Sivic, Tomas Pajdla, and Akihiko Torii.
\newblock Inloc: Indoor visual localization with dense matching and view synthesis.
\newblock In {\em Proceedings of the IEEE conference on computer vision and pattern recognition}, pages 7199--7209, 2018.

\bibitem{tardioli2015visual}
Danilo Tardioli, Eduardo Montijano, and Alejandro~R Mosteo.
\newblock Visual data association in narrow-bandwidth networks.
\newblock In {\em 2015 IEEE/RSJ International Conference on Intelligent Robots and Systems (IROS)}, pages 2572--2577. IEEE, 2015.

\bibitem{wang2022matchformer}
Qing Wang, Jiaming Zhang, Kailun Yang, Kunyu Peng, and Rainer Stiefelhagen.
\newblock Matchformer: Interleaving attention in transformers for feature matching.
\newblock In {\em Proceedings of the Asian Conference on Computer Vision}, pages 2746--2762, 2022.

\bibitem{yi2016lift}
Kwang~Moo Yi, Eduard Trulls, Vincent Lepetit, and Pascal Fua.
\newblock Lift: Learned invariant feature transform.
\newblock In {\em Computer Vision--ECCV 2016: 14th European Conference, Amsterdam, The Netherlands, October 11-14, 2016, Proceedings, Part VI 14}, pages 467--483. Springer, 2016.

\bibitem{zhang2018learning}
Linguang Zhang and Szymon Rusinkiewicz.
\newblock Learning to detect features in texture images.
\newblock In {\em Proceedings of the IEEE conference on computer vision and pattern recognition}, pages 6325--6333, 2018.

\bibitem{zhou2016evaluating}
Hao Zhou, Torsten Sattler, and David~W Jacobs.
\newblock Evaluating local features for day-night matching.
\newblock In {\em Computer Vision--ECCV 2016 Workshops: Amsterdam, The Netherlands, October 8-10 and 15-16, 2016, Proceedings, Part III 14}, pages 724--736. Springer, 2016.

\end{thebibliography}
}

\end{document}